# Semantic Contextual Reasoning to Provide Human Behavior

Sarika Jain and Archana Patel*
National Institute of Technology Kurukshetra, India
Software Engineering Department, Eastern International University, Vietnam
jasarika@nitkkr.ac.in, archanamca92@gmail.com

**Abstract:** In recent years, the world has witnessed various primitives pertaining to the complexity of human behavior. Identifying an event in the presence of insufficient, incomplete, or tentative premises along with the constraints on resources such as time, data and memory is a vital aspect of an intelligent system. Data explosion presents one of the most challenging research issues for the intelligent systems; to optimally represent and store this heterogeneous and voluminous data semantically in order to provide human behavior. There is a requirement of intelligent but personalized human behavior subject to constraints on resources and priority of the user. Knowledge, when represented in the form of ontology, procures intelligent response to a query posed by users; but it does not offer content in accordance with the user context. To this aim, we propose a model to quantify the user context and provide semantic contextual reasoning. A diagnostic belief algorithm (DBA) is also presented that identifies a given event and also computes the confidence of the decision as a function of available resources, premises, exceptions, and desired specificity. We conduct an empirical study in the domain of day-to-day routine queries and the experimental results show that the answer to queries and also its confidence varies with user context.

## 1. Introduction

Humans utilize their common sense, their life long experiences, creativity, innovation and intelligence to make decisions. With an Artificial Intelligence (AI) perspective, we can train computers to outperform humans in many tasks. Since the advancement in AI technologies, humans have been reconfiguring their work practices in their day-to-day lives starting from their household activities to office and business, making their lives convenient and comfortable. AI technologies disguise the human labor by integrating it into social context, hence profoundly and dynamically changing the daily routines. Simulating general intelligence comprises of traits like representing knowledge, reasoning, planning and learning. Knowledge Engineering has been recognized as a powerful key for building large scale AI applications and is being looked as the key for neuro-symbolic computing. AI researchers started to use the term ontology as a component of knowledge-based systems from 1980s [1]. Ontological dimensions of AI are exclusively under investigation today and are reshaping the modern systems. In addition to clever algorithms, rich knowledge bases acquainted with context information plays an intrinsic role in achieving specific expert performance. Recently a wave has emerged to enable the automatic identification of human activity / event / emotions / location / situation collaboratively called as human behavior. In real-world different types of users communicate with the same system. The same query posed to a user may generate different answers depending upon the variable factors (available resources, allowed certainty, allowed specificity and acceptable threshold) i.e. the context of the user. It is required for the system to behave according to the context of the users. Human context ontologies provide personalized behavior that involves the design of enabling systems to capture or infer the needs of each person and then to satisfy those needs in a known context. Various human context ontologies like 3LConOnt, PiVOn, and mIO! have already been developed to recognize the behavior of the human. These contextual semantic data models for event identification, prediction, and exploitation provide the following benefits:

- Event data becomes meaningful information.
- Better understanding of situations (states) by machines.
- Better understanding of the relationships between events.
- Declarative processing of events and reaction to situations.
- Powerful semantic querying to build personalized human behavior
- Ability to recognize current context increases usability and effectiveness

Despite all this, some challenges still remain. Decision making is made simple if the coded intelligence is actionable. If a system does not behave according to the user context, then achieving results may not be sensible. An Agent in the real environment is inevitably forced to make a decision with incomplete information. In the real world, the domain knowledge as well as the contextual information are somewhat imperfect, meaning that the imprecision and uncertainty are not represented; yet demanding for a personalized smart response. The smart response, if provided, develops a synergy between the different answers to the same query posed by users. The existing human context ontologies miss to represent the features of uncertainty, vagueness, and imprecision [2]. The same can be achieved if we have a good knowledge representation scheme that enables us to represent every entity as a whole, i.e., with complete information. Followed by Jain and Jain [3], Patel et al. [4] have presented such a knowledge representation scheme that is detailed further in following sections. This knowledge representation scheme provides us with a unit of knowledge that is further exploited in this paper to quantify the user context.

An event or activity is an outcome or a trigger or an occurrence which carries some significance. Several reasoning architectures and prototypes have been proposed for event detection and even for event recognition, each of which follows a different approach for handling characteristics of the domain. The event identification facilitates various services like situation awareness / information dissemination, strategic and operational decision making, teaching and training, timely response to emergencies or business decision, and real-time surveillance. However, identification of the event according to the user context is a challenging task because a user has different constraints on available resources, certainty, specificity, belief, and preferences. This makes up the variation in the confidence of the results/outcomes. Say it is asked, "Where is Robert?" The answer may be

- Robert is in city Mumbai, if asked to be very quick.
- Robert is working outdoor in city Mumbai if asked to be specific, may be less certain.
- Robert is planting crops in city Mumbai, if asked to be highly specific.

Intelligent systems may be based on variable certainty or variable specificity or both of them. In variable certainty systems, the more is the given time, the more certain are the answers. In variable specificity systems, the more is the given time, the more specific are the answers. Both the approaches should be combined to reflect the fact that specificity and certainty are the two opposite sides of a coin. One has to be sacrificed to achieve the other. The major challenge is to maintain the tradeoff 'certainty vs. specificity' and provide reasoning under resource constraints with uncertain, imprecise (event is correct, yet inexact), and inaccurate (event is completely or partially wrong) information.

In this paper we have devised a model to express the context in different dimensions and using several parameters. We exploit the semantic technologies to model the personalized behavior and hence provide semantic contextual reasoning. The proposed approach recognizes the pattern in the flow of confidence as the priority of user changes. A control protocol has been devised to define the priority of the user and is exploited in an efficient algorithm to yield good tradeoffs between various attributes of the decision. The algorithm renders varying answer with varying confidence based on a balance between the available resources, the required certainty, the required specificity level, and the acceptable threshold value. The granularity also plays an intrinsic role; the finer the details of context, the better the confidence in decision provided. The aim of this paper is to provide semantic contextual reasoning by virtue of recognizing an activity/event semantically and contextually with a confidence value. In addition, to provide answers to queries, the proposed work also quantifies the quality and depth of the answer. Our contributions in this paper are as follows:

- To design a human context ontology in order to provide human behavior along with its evaluation with existing human context ontologies.
- To propose a model and control protocol for quantifying the context and device a semantic contextual reasoning approach for identifying a given event/activity.
- To provide a simulation prototype to answer daily life queries with a confidence score.

The remainder of this chapter proceeds as follows: section 2 presents a consolidated literature review about the existing human context ontologies and their comparison on essential features. Section 3 designs a normal routine ontology to answer daily life queries after first describing the knowledge representation structure utilized. The proposed algorithm for semantic contexual reasoning along with the control protocol for quantifying the user context has been featured in section 4 and a comprehensive evaluation of the proposed work is presented in section 5. Section 6 concludes the paper with some light on future plans.

## 2. Human Context Ontologies

The study of ontologies is based in the works of Plato and Aristotle. An ontology represents the hierarchical categorization of entities. Databases represent and store data and information while knowledge organization systems (KOS) are systematized knowledge of domains and make search more robust. The simplest type of KOSs is the controlled vocabulary, then taxonomies, thesauri, folksonomies, and ontology getting better. Many evaluation frameworks exist with different criteria and approaches including licensing, structure, purpose, maturity, granularity, autonomy, intelligence, vagueness, integration, security, service availability, and more. For contextual reasoning and human activity recognition, we need to model human behavior in ontologies.

## 2.1. Ontologies Modeling Context

An ontology offers good semantics that helps to provide an accurate and precise answer. A context is information that is used to describe the situation of an entity, where an entity can be a place, object, and person [5]. Human context ontology plays a very important role during personalization because it depicts the behavior of reasoning and provides precise answer according to the choice of the user. Keeping this vision, several researchers have utilized ontologies to recognize context of the user. There are broad varieties of ontologies available to model the context in smart environments. Here we list various domains/variables/features that we look for during ontology modeling in order for it to qualify as human context ontology [6] and Table 1 provides a description of existing human context ontologies.

- User, Role, Location, Environment, Time, and Granularity are the most basic parameters. The 'User' element model the information about the user's profile and preferences. 'Role' of the user determines his facet under consideration. The 'Location' element describes the location of the context, for example, indoor activity or outdoor activity. 'Environment' element model information about the environment like humidity. 'Time' element models temporal knowledge and timestamp of event. A user may have different role and context at different times of the day. 'Granularity' levels of behavior/action/task/activity. Behavior granularity explains the depth of the answer means how much answer is specific
- Taxonomies of interface, network, device, provider, and service are required to after the human activity has been recognized. 'Interface' feature is available for input/output functions. 'Network' feature models the specification of the network. 'Device' element models the knowledge about devices like networking device, and computer. The element 'provider' provides the services according to the environment. 'Service' element models the knowledge about the type of services offered.
- 'Context Sources' is essential to know the origin of the contextual information. The 'Server' elements within the ontology act as Context Source like the user, device, role, environment and so on
- 'Message' element formalizes the message.
- 'Social interaction', both virtual (for human-to-computer messages) and physical (for human-to-human messages) involves the exchange and formalization of the message with feedback from the users.
- 'Implementation available' element checks that implementation is available or not.
- 'Imprecision, Vagueness, and Uncertainty management' provide a certain value of the answer.
- 'time' is a standardized common representation of universal entities, such as time and geographical indoor and outdoor locations, as well as environmental conditions
- relations such as ownership of objects, rights, services, or privacy and service access

Table 1. Description of Existing Human Context Ontologies

| Human Context Ontology | Description |
|---|---|
| CC/PP [7] | specially designed for wireless devices that allow the server to convey content. Device sends CC/PP profile (which constructed as a 2-level hierarchy) during making a request by using protocol. CC/PP is based on RDF and XML format. An extension of CC/PP ontology is proposed by [8]. The main aim of the authors is to improve interoperability by providing a static and dynamic characteristic of devices along with proper conditions that trigger pre-defined adaptations. |
| COBRA-Ont [9] | contains a collection of ontologies for describing agents, environment, places and their associated properties in the domain of intelligent meeting room. This ontology is expressed in OWL and contains 41 classes and 36 properties. A root class of COBRA-Ont is "place" that has set of properties to describe a location. |
| CoDAMoS [10] | designed to solve the challenges of application adaptation, code mobility, automatic code generation, and user interface according to the specific devices. It contains 4 main concepts namely user, platform, service, and environment. |
| Delivery context ontology [11] | models the characteristic of the environment where different devices interact with services. The main features of deliver context ontology are hardware, software, environment, and location. |
| SOUPA [12] | expressed in OWL syntax and referenced various ontology like FOAF, spatial ontologies in OpenCyc, COBRA-Ont and so on. It is divided into two blocks namely SOUPA-core and SOUPA-extension and includes modular component vocabularies. |
| mIO! Ontology [6] | developed by using NeOn methodology and defines by 11 modular ontologies. It is a network ontology that provides services according to user interest. |
| PalSPOT ontology [13] | maps context data to ontological classes and properties. It uses hybrid ontological reasoners or statistical reasoners to retrieve the information about human activity |
| CONON [14] | encoded in OWL and defines general concepts like location, activity, person and so on. CNON adds domain-specific entity in a hierarchal way and model different intelligent environment. It also includes domain-specific ontologies like |
| | home domain and office domain ontologies. |
| PiVOn ontology [15] | composed of 4 ontologies namely user, environment, devices and services which provides an intelligent environment. 5Ws theory is used to describe the context. |
| Situation ontology [16] | divided into two layers namely situation and context layers. Situation layer defines applications as a set of context whereas context layer defines user, environment and device context. An entity is satisfied with the situation according to the context data that have a certain value. |
| CAMeOnto ontology [17] | a meta ontology for context and extends the specific information of the context in a hierarchal manner. It follows 5Ws theory along with six conceptual classes namely user, time, service, activity, location and device. |
| 3LConOnt [18] | a 3-level context ontology namely upper-level ontology, middle-level ontology, and lower level ontology. It can be easily reused, extended and adapted for different purposes. |
| Sup_Ont [19-20] | an upper ontology based on the structure of EHCPR. In this ontology the structure of universe shows the concept of reality that is defined to have an existence which is known as truth. Malik and Jain [21] have also developed a context model based on the scheme of the EHCPR. |

Rodríguez et al. [2] have compared diff Human Context Ontologies and ranked them semantically based upon the number of parameters of the context handled by them. They have also compared them on structural point of view using the OOPs Pitfall scanner for determining anomalies or errors. Table 2 presents a comprehensive summary of the evaluation over the existing human context ontologies.

**Table 2. Comparative Analysis of Existing Human Context Ontologies**

| Ontology Features | CC/PP | CoBrA-Ont | CoDAMoS | Delivery Context | SOUPA | mIO! | PalSPOT | CONON | PiVOn | Situation Ontology | CAMeOnto | 3LConOnt | Sup_Ont |
|---|---|---|---|---|---|---|---|---|---|---|---|---|---|
| Device | √ | √ | √ | √ | | √ | √ | √ | √ | √ | √ | √ | √ |
| Environment | | √ | √ | √ | | √ | √ | √ | √ | √ | √ | √ | √ |
| Interface | | | | √ | | √ | | | | | | √ | √ |
| Location | | √ | √ | √ | √ | √ | √ | √ | √ | √ | √ | √ | √ |
| Network | √ | | | √ | | √ | | √ | | | | √ | √ |
| Provider | | | | | | √ | | | √ | | | √ | √ |
| Role | | √ | √ | | | √ | | | √ | | √ | √ | √ |
| Service | | | √ | √ | | √ | | √ | √ | | √ | √ | √ |
| Context Source | | | | | | √ | √ | | √ | √ | | √ | √ |
| Time | | √ | √ | | √ | √ | √ | √ | √ | √ | √ | √ | √ |
| User | √ | √ | √ | | √ | √ | √ | √ | √ | √ | √ | √ | √ |
| Uncertainty Management | | | | | | | | | | | | | |
| Message | | | | | | | | | √ | | | √ | √ |
| Behavior Granularity | | √ | √ | | | | √ | √ | √ | √ | | | √ |
| Social Interaction | | | | | | | √ | | √ | | | | √ |
| Implementation available | √ | √ | √ | | √ | √ | √ | | | | √ | √ | √ |

It can be noticed that the human context ontologies still require to model imprecise and uncertain information to represent the real life scenarios and activities. The formalization of the message is required to be modeled as it provides feedback from the users and a way of interaction to the system. Social interaction provides message exchange from the system to the user and vice-versa. All these features should be incorporated in the human-context ontologies. Otherwise, we need to map various heterogeneous ontologies in order to inherit all the listed parameters from some ontology or the other.

**2.2. Encoding of Contextual Parameters in Ontology**

Web ontology language (OWL) is the most widely used ontology language [22]. Several extensions of OWL have been proposed by various authors because it does not deal with uncertain, imprecise, temporal and spatial information. To play with uncertainty, various mathematical frameworks for extension of OWL such as fuzzy, possibilistic and probabilistic extensions with DL formalism have been reported in the literature. Fernando and Umberto [23] explained that there are two ways to deal with vague information in OWL. First one is to extend the current language and the second one is to provide the procedures to represent such information within the language.

They follow the second approach and proposed a methodology to build fuzzy ontology by using annotation property of OWL2. Stoilos et al. [24] proposed fuzzy OWL to capture imprecise and vague knowledge. They presented a reasoning platform called a fuzzy reasoning engine. Bobillo et al. [25] proposed a methodology to represent uncertain knowledge in ontology called fuzzy ontology using OWL2 annotation properties. They presented a plug-in for development of fuzzy ontology. Viorel et al. [26] have presented a temporal extension of OWL called temporal OWL which is expressive fragment (SHIN(D)) of DL. They used layered approach and introduced three extensions in OWL namely concrete domain (allow to present restriction), temporal representation (introduce time points and its relation, intervals), fluents/timeslices (implement perdurantist view and complex temporal aspects). Lee et al. [27] have proposed type2 fuzzy ontology model (T2FO) which rooted on interval type-2 fuzzy sets. They applied the proposed model for the representation of knowledge in the domain of diabetic-diet recommendation. The T2FO consists of T2F personal ontology, T2F food ontology, and T2F personal food ontology. Ausin et al. [28] presented an approach called TURAMBAR to interfuse OWL2 reasoning and expressibility power by using Bayesian networks to overcome the well-known limitations of OWL. Stoilos et al. [29] have proposed a fuzzy extension of OWL called f-OWL and presented a translation method which reduces the inference problem of f-OWL into inference problems of fuzzy DLs. OWL DL has expressiveness limitations as it does not support temporal reasoning that is crucial for capturing complex temporal relations between activities.

Production rules (if-then) are another way to represent the dynamic information in such a way that provides quantification of personalized behavior. Michalski and Winston [30] have added unless operator in production rule called CPR as a primordial computational and representational scheme for Variable Precision Logic (VPL) in which specificity remains constant and only certainty varies. Hewahi [31] has presented a rule structure called Concept Based Censor Production Rule (CBCPR), an extension of CPR. Each rule has a certain concept title that specifies its job. He claimed that this structure of rule will help the system to provide more certain answers within a specified time. Bharadwaj and Jain [32] have presented HCPR (If, Unless, Generality, Specificity operators) as a knowledge representation scheme. HCPR deals with certainty as well as specificity, two main facets of precision. They have shown the trade-off between the computational efficiency and precision of inference in HCPRs system and used various control parameters for quantization of the answer. Later on, Jain and Jain [33] presented an extension of HCPR called EHCPR where properties are divided into two categories namely defining and characteristic property. Jain et al. [34] described reasoning in the EHCPRs system along with the importance of default and constraints list of concept. They mentioned that EHCPRs is a system which provides reasoning with real-life problems. Jain and Jain [35] represented the structure of constraints and defaults in EHCPRs system. They described seven conditions where constraints have to be imposed on the system. They separated knowledge base into declarative knowledge and procedural knowledge. Patel et al. [4] have encoded the idea of EHCPRs into ontology; hence describing real-world entities as a knowledge unit by utilizing existing constructs. They have provided domain ontology of emergency situations where every node of the tree contains complete information. The proposed knowledge unit is encoded in RDF/XML format of OWL2.

## 3. Normal Routine Context (NRC) Ontology
The contextually rich data sets suitable for modeling the personalized behavior are not readily available. We have modeled the domain of daily life queries as a normal routine context (NRC) ontology and curated it manually. The user will be able to perform reasoning over NRC. The NRC ontology considers the problem statement of the day-to-day routine tasks in which it is required to know about what a particular person is doing at some instant of time.

### 3.1 Formalization of Entity
Several ontology languages are available in the literature such as DAML+OIL, OIL, SHOE, RDF, RDFs and OWL, but OWL is most widely used ontology language. The OWL ontology is a knowledge representation formalism based on description logic to represent real-world entities according to the semantic data model. This type of representation does not deal with the problem of vastness, vagueness, uncertainty, and inconsistency. Classical logic of ontology permits conclusions which are either true or false. It does not permit the degree of truthfulness. However, in the real world, the queries posed by the user may have variable answers with variable confidence. For instance, if a group of users is asked to identify an animal, there is a diversity of perspectives and options as they reason with constraints on resources of time, their memory, their knowledge, uncertainty of knowledge and requirements such that (i) highly certain answer may it be less specific (ii) Highly specific answer may it be less certain. Mapping the diversity of these answers into a meaningful spectrum ignoring some fact or the other based on the available resources motivates us to define the precision of the answer based on the tradeoff between available resources, allowed certainty, allowed specificity and acceptable threshold. Using the structure given by Patel et al. [4], let us represent every concept in ontology in the form of the following tuple:

$$D\,[TE, AE, VE, PE](\omega) \;=\; <DF(\gamma), CF, C(\delta), G, S, I>$$

| | | |
|---|---|---|
| D [TE, AE, VE, PE] ω | {Decision/Concept/Event} | |
| If$_{dis}$ | DF{DF$_1$:ω$_1$; DF$_2$:ω$_2$; ... DF$_m$:ω$_m$} γ | {Distinctive Features (ANDing)} |
| If$_{can}$ | CF {Pa$_1$:P$_1$C$_1$; Pa$_2$:P$_2$C$_3$;...,Pa$_p$:P$_3$C$_1$} | {Cancellable Features} |
| Gen | G | {General concept} |
| Spec | S {S$_1$, S$_2$, ..., S$_k$} | {Specific concepts (XORing)} |
| Unless | C {C$_1$:δ$_1$; C$_2$:δ$_2$; ...; C$_n$:δ$_n$, UNK}, δ | {Exceptions to the rule (ORing)} |
| Instances | I [Temporal Details, Spatial Details] | {Instances} |

**Fig 1. Concept in Ontology**

The If part of the rule has two types of features, the distinctive and the cancellable features. DF operator as relegated with the If$_{dis}$ part of the rule enumerates the set of m preconditions/Distinctive Features (If condition), which should always be satisfied to draw the decision D, which is the THEN part of the rule. CF operator as relegated with the If$_{can}$ part of the rule enumerates the set of p Cancellable Features associated with the concept D. Cancellable features of a concept in the hierarchy may or may not be derived. These features can be overridden or new features can be added at each concept in the taxonomy. These features are usually held for an instance but it is not mandatory. Both the distinctive and cancellable features are further of two types: physical parts and abstract properties. P$_1$C$_1$ is the default value of part Pa$_1$ for the concept D chosen from the constraints list of part Pa$_1$ {P$_1$C$_1$, P$_1$C$_2$...}.

G operator refers to the generality part of the rule. It is the general concept of concept D. S operator refers to the specificity part of the rule. It enlists the set of k concepts specific to the concept D in the taxonomy of ontology.

C operator as relegated with the UNLESS part of the rule enumerates the set of n exceptions to the rule. Every exception has a probability value denoted by $\delta_1, \delta_2 \ldots \delta_n$. The Unless operator refers to the list of censors.

I is the list of known individuals/instances of the concept D and is relegated with the Instances part of the rule.

Every premise (DF$_1$, DF$_2$,....DF$_m$) is associated with a probability value ($\omega_1, \omega_2, \ldots \omega_m$) which shows the confidence of the truthfulness of DF. If all values of premises are greater than or equal to the assigned value of threshold (DF$_{Thres}$) then only that node will be considered for decision. TE, AE, VE, and PE, are respectively the textual encryption, audio encryption, video encryption, and pictorial encryption of the concept D. Factor $\gamma$ is the precision of the decision with 0-degree of strength (if-then relationship), factor $\delta$ is the precision of the decision with 1-degree of strength (also considering exceptions) and factor $\omega$ is the precision of the decision with 2-degree of strength (also considering hierarchy). If an exception is evaluated to be true, it will negate the decision completely. An unknown exception will contribute 0 as its $\delta$ value and an exception evaluated to be false will contribute its $\delta$ value to the decision according to equation 1. The values of $\gamma$ and all $\delta$'s are constant.

$$\delta = \gamma + Summation\ of\ all\ \delta's \tag{1}$$

For useful implication, the value of $\delta$ should lie between $0.5 < \delta \leq 1$. The precision of the decision with 2-degree of strength of node D at some level i is calculated as:

$$\omega_D(i) = \min(\omega_D(i-1), \omega_{P_1}, \omega_{P_2}, \ldots \omega_{P_m}) \times \delta_i \tag{2}$$

Where, $\omega_D(i-1)$ is the precision of parent class with 2-degree of strength. We always consider $\omega_D(i-1) = 1$ when calculating $\omega_D$ of the root node. To cater incomplete information, we have used UNK that represents a disjunction of unknown censors. This type of representation reduces the existing challenges of knowledge representation and leads us to forms of representation that we believe are more natural and comprehensible than other forms. We have used annotation properties of OWL for storing the defining and cancellable properties and the $\gamma$ and $\delta$ parameters of the entity. For storage of exceptions, we have utilized the disjoint relationship. The textual encryption of the entities is stored by using the label construct, whereas the remaining encryptions are stored by using data properties. The constraints list have been provided by OWL:oneOf construct.

All these attributes and parameters form an inseparable cluster, i.e., a unit of knowledge, which together help in quantifying the user context and provide the personalized behavior. To model the context it is required to encode complete information as a unit in every concept of ontology. This knowledge unit encodes every concept as a whole and can be utilized in various domains; for example in the domain of emergency, user can ask

'what is the event that happed at Place L on date M' and in domain of daily life queries, user can ask 'where is person P?', 'what is Person P doing?' and so.

### 3.2 Engineering the NRC

Based on the idea of knowledge unit as discussed in section 3.1, we have curated Normal Routine Context (NRC) Ontology. Upper ontologies support broad semantic interoperability among domain ontologies by providing a common platform for the formulation of the definitions. The domain ontology NRC is integrated with upper ontology SupOnt. SupOnt ontology is an upper ontology that allows to model any environment [20]. SupOnt ontology contains all the top-level features that are required to represent the context. Figure 2 depicts the taxonomy of SupOnt-NRC ontology for the normal routine tasks, a person might be performing. The upper part of the figure shows few concepts of SupOnt ontology and lower part of the figure displays the complete NRC ontology. Every concept carries its complete knowledge as a unit. NRC ontology is just indicative of what P, a person may be doing at any instant of time. It is created as a sample to diagnose the confidence of the decision. If person P has DFs "lives in city R" and "works in city R", then that P is found in city R unless P is on tour or has long vacation or is entertaining outside the city. The event 'isInCityR' is divided into two sub-events namely 'isAtHome' and 'isOutdoor' where 'isWorkingOutdoor' and 'isEntertainingOutdoor' are the two subevents of 'isOutdoor'. The event 'isWorkingOutdoor' is further classified into many events. The node 'Other' caters to rest of tasks P may be doing while working outdoor. The total number of levels in NRC ontology is currently 4. The $\gamma$ and $\delta$ values have been provided based on the authors' experience and are subject to change on case-to-case basis. The $\delta$ value of UNK has been taken to be 0.01 consistently. The NRC ontology is connected with the SupOnt ontology by subClassOf relationships as mentioned in table 3. A case study over SupOnt-NRC ontology is taken for the diagnosis of daily life queries to explain the mathematical simulation of semantic contextual reasoning.

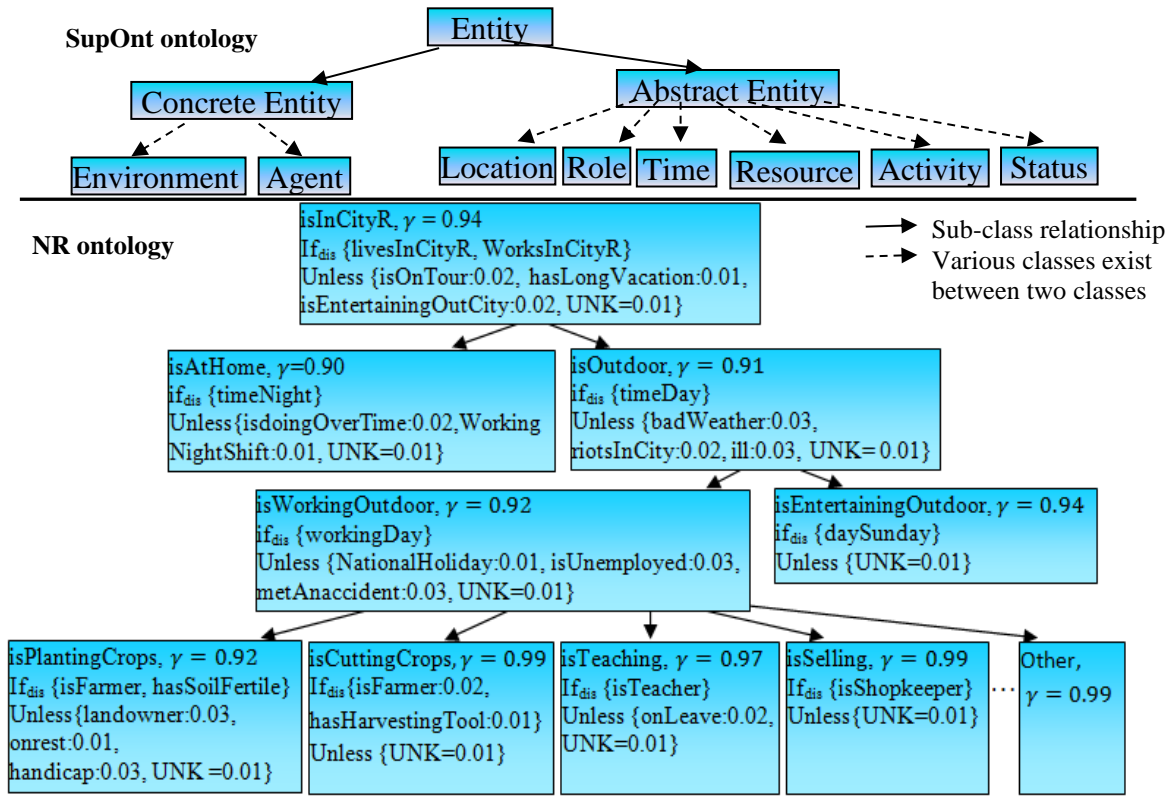

**Fig 2. Representation of complete knowledge for diagnosis of normal routine queries**

Table 3. Integration of NRC ontology with SupOnt ontology

| SupOnt | NRC |
|---|---|
| Environment | badWeather |
| Agent | isFarmer, isTeacher, isShopkeeper |
| Location | isAtHome, isOutdoor, isWorkingOutdoor, isEntertainingOutdoor |
| Time | timeNight, timeDay, workingDay, daySunday |
| Resource | hasHarvestingTool |
| Activity | isInCityR, livesInCityR, WorksInCityR, isOnTour, hasLongVacation, isEntertainingOutCity, isDoingOverTime, WorkingNightShift, riotsInCity, metAnAccident, isPlantingCrops, isCuttingCrops, isTeaching, isSelling |
| Status | ill, isUnemployed, landOwner, onRest, handicap, onLeave, hasSoilFertile |

## 4. Semantic Contextual Reasoning

To provide personalized behavior according to the context is a vital aspect of the intelligent system. Several information systems have been proposed in order to personalize information or provide more relevant and appropriate information for users. Zhang et al. [36] have described context-aware framework based on an ontology for behavior analysis that addresses the problem of efficient and appropriate delivery of feedback by combining context to manage any inconsistency between what the user is expected to do and what the user is actually doing. Calegari and Pasi [37] have improved the quality of web search by using personalized ontologies. They applied two techniques namely re-ranking and query reformation and defined the observation of information behavior that helps the user in finding relevant information. Niaraki and Kim [38] have designed a generic ontology-based architecture for personalized route planning system by using a multi-criteria decision-making technique. Mittal et al. [39] have proposed a hybrid approach by using an ontology, user profile and collaborative filtering techniques for personalized web Information Retrieval. They provided an empirical analysis of proposed work that shows the improvement over precision, recall, and f-score. Nafea et at. [40] have created a personalized environment that involves arrangement of information about each learner. They used an ontology, Myers-Briggs Type Indicator theory (MBTI), rule-based reasoning and Felder-Silverman learning style model (FSLSM) for the creation of personalized student profile.

A user context is a vague term which incorporates user's thoughts and expectations from the system. There is a need to quantify the user's expectations for the thresholds, the requirement of certainty viz-a-viz specificity and also the resources available to him to perform the reasoning. This all makes up what is called the context of the user. Figure 3 shows varying user context during the event or activity identification. The proposed semantic contextual reasoning uses control protocol and diagnostic belief algorithm (DBA) to identify the event with 2-degree of confidence. It takes input either from the event stream or event query and fetches the information of the event from the knowledge base. The event is preceded according to the value of contextual information (CI). The output of the algorithm can be utilized in various applications like decision making, situation awareness, real time surveillance and timely response to emergencies.

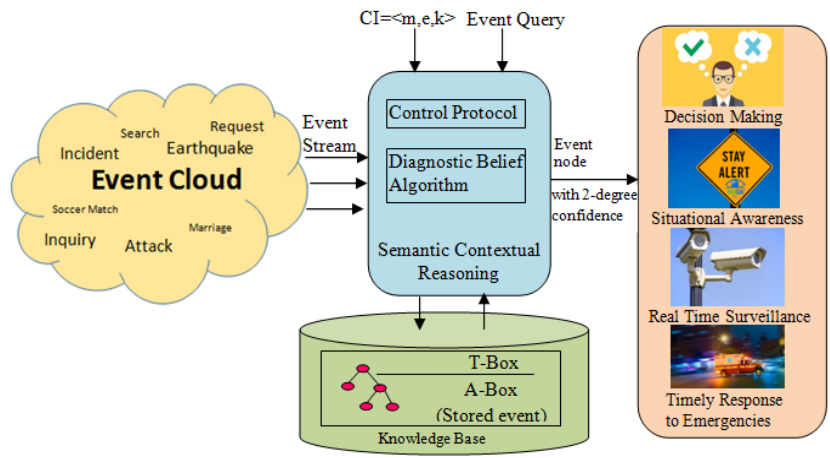

**Fig 3. Event Identification according to the user context**

## 4.1. Control Protocol

In order to maintain a balance between the user expectations and the system generated answer, we represent context in two dimensions and device a control protocol for the same. 1st dimension explains which event will be the answer according to the Distinctive Features (DFs) and 2nd dimension specifies how much the answer will be specific according to the type of the user. The control protocol provides procedures to handle trade-offs between the confidence of inferences and computational efficiency of deriving them and also facilitates trade-off between the certainty of the conclusion and its specificity. The control protocol to be followed by the reasoning process will depend upon two control parameters; the threshold of premises ($DF_{Thres}$) and contextual information (CI). Values to these parameters would be assigned according to the various resource constraints and requirements imposed by the user, i.e., his priority level. These control parameters are defined as follows:

1. **Threshold of premises ($DF_{Thres}$):** If the probability value of all premises of a concept is greater than or equal to $DF_{Thres}$, the decision will be considered true otherwise false. The default value of $DF_{Thres}$ is set to 0.88.
2. **Contextual information (CI):** The contextual information associated with each user has been defined by a set of 3 attributes:

$$CI = <m, e, k> \quad (3)$$

where

- m represents the threshold confidence in decision. The answers with confidence in decision ($\omega$) $\geq m$ would be considered true and answers with confidence in the decision ($\omega$) $< m$ would be considered false. The value of the parameter m lies in the range (0, 1] and the default value is 0.49.
- e represents the available resources. The parameter e informs about total available resources. The different sample ranges of parameter e for different users are mentioned in Table 4(a). These different ranges of parameter e may vary according to the type of knowledge in the knowledge-base and the performance of the reasoning system. The default value of the parameter e might be 1.0 indicating that the reasoning system normally has some resources like time or memory or both.
- k represents the tradeoff between certainty and specificity. The parameter k allocates the available resources according to the choice of users for certainty vs. specificity. For demonstration purpose, we have given different ranges of k that provides an answer to the imposed query with respect to the user requirement over specificity vs. certainty. The range of parameter k is listed in Table 4(b).

The value of CI = <m, e, k> has been utilized to define the priority of the user and governs the granularity of the decision. The range of m for different value of e and k has been defined for five different types of users as in table 4(c).

**Table 4. Quantifying the User Context (a) Available Resources (b) Certainty vs. Specificity (c) Defining User Priority**

(a)

| Resource availability | Range of parameter 'e' |
|---|---|
| Very-2 low | [0.125, 0.25) |
| Very low | [0.25, 0.50) |
| Low | [0.50, 1.0) |
| Moderate | 1.0 |
| High | (1.0, 2] |
| Very high | (2.0, 4.0] |
| Very-2 high | (4.0, 8.0] |

(b)

| User requirement on Certainty | User requirement on Specificity | Range of parameter 'k' |
|---|---|---|
| Very high | very low | [0.25, 0.50) |
| High | low | [0.50, 1.0) |
| Moderate | moderate | 1.0 |
| Low | high | (1.0, 2.0] |
| Very low | very high | (2.0, 4.0] |

(c)

| Type of User | m | e | k |
|---|---|---|---|
| Very Low Priority (VLP) | [0.3, 0.4) | [0.125, 0.25, 0.4] | [0.25, 0.5, 1, 2, 4] |
| Low Priority (LP) | [0.4, 0.45) | [0.5, 0.6, 0.8] | [0.25, 0.5, 1, 2, 4] |
| Moderate Priority (MP) | [0.45, 0.55) | [1, 1.2, 1.5] | [0.25, 0.5, 1, 2, 4] |
| High Priority (HP) | [0.55, 0.7) | [2, 3, 4] | [0.25, 0.5, 1, 2, 4] |
| Very High Priority (VHP) | [0.7, 0.8] | [5, 6, 7, 8] | [0.25, 0.5, 1, 2, 4] |

The user has to specify the context in the form of tuple $CI = <m, e, k>$ each time, he queries the system. The maximum level of specificity (N) and the maximum level of certainty (X) of an answer are calculated according to the following formulas:

$$Max. Specificity\ (N) = f1\ (m, e, k, N_m)$$

i.e. $$N = ceiling\left(N_m * \left((1.0 - m) \wedge \left(\frac{1}{e*k}\right)\right)\right) \quad (4)$$

$$Max. Certanity\ (X) = f2(m, e, k, N, CL, X_m)$$

i.e. $$X = Round\left(X_m * \left(1.0 - (1.0 - m) * \left(\left(\frac{CL+1}{N}\right) \wedge \left(\frac{e}{k}\right)\right)\right)\right) \quad 0 \leq CL < N \quad (5)$$

where, $N_m$ is the total number of levels (including level-0) in the taxonomy of ontology corresponding to which the query is issued, and N is the maximum number of levels of specificity of ontology taxonomy (including level-0) to be explored under a particular context. CL denotes the current level of the running node, $X_m$ is the total number of exceptions at that node, and X is the maximum number of exceptions (maximum number of certainty) that will be computed out of $X_m$ at level CL.

**4.2. Diagnostic Belief Algorithm**
Here we propose a diagnostic belief algorithm (DBA) that quantitatively represents the expectation of user by virtue of the control protocol and precisely predicts the answer. The available resources, allowed certainty, allowed specificity, and acceptable threshold are the variable factors determining the context of the user. DBA is a step by step procedure of diagnosis of the event using a combination of four set of parameters {$DF_1$, $DF_2$…$DF_m$}, {$X_1$, $X_2$…$X_n$}, CI = <m,e,k>, and $DF_{Thres}$ to identify the confidence of the answer. Following mentioned pseudo-code computes the confidence of decision. Initially, DBA sets the current level (CL) equal to 1 and fetches the total number of the levels ($N_m$) in the taxonomy of ontology then computes the maximum number of levels (N) by function calcN() that needs to be traversed. It picks the root node and insert it into the queue (we used queue data structure so that node will be processed according to the sequence of First-In-First-Out). Now, remove one node (cNode) from the queue and check the premises of that node according to the function checkPremise(). Ontology tree is traversed according to breath first search but when the premises of a node satisfies (use function checkPremise()) then at that level no other siblings need to be explored. When the premises of a node are satisfied, calculate a number of exceptions (use function calcX()) that need to be checked at that node. If any exception is true at a node then answer (use function calcω()) will be given by parent node of that node. The queue data structure is used to store all the nodes that need to be checked whereas stack data structure is used to store all processed parent nodes. We have set the default value of m to be 0.49.

**Soundness, Completeness and Termination:** To check the soundness of the algorithm, we have run approximate 100 cases over the DBA algorithm and every test case always returned correct answer of the imposed queries according to the constraints of the users. Some selected test cases with different combinations of parameters are discussed in section 5.3.1 showing the soundness of the DBA algorithm.

checkPremise() is the first function which is responsible to return always correct and precise answer according to the input value of the users because this function allows us to go down the hierarchy only if the value of all the premises of the node is equal or above the $DF_{thres}$. By using the formula calcδ(), we check the status of the exceptions. If any exception is true at any node then that node will not contribute in the decision and answer will be given by the parent node. calcδ() handles all the exceptions and display the correct and precise decision with 1-degree of strength. The value of parameter m limits/restrict the confidence of the decision with 2-degree of strength that are calculated by the function calcω(). The answer with confidence in decision ($\omega$) $\geq m$ would be considered true and answers with confidence in the decision($\omega$) $< m$ would be considered false. These three functions (checkPremise(), calcδ() *and* calcω()) prove the completeness of the algorithm because they incorporate all the exceptions and guarantees to return a correct answer for any arbitrary input (or, if no answer exists, it guarantees to return failure) and any condition.

The DBA algorithm can be terminated at three points, line number 5, 9 and 13. Section 5.3.1 shows all the possible cases for the termination of DBA algorithm for every type of user (VLP, LP, MP, HP and VHP). For example, in case of VLP user, Case 1 is terminated because of CL≤ N, Case 2 is terminated because of the true of value exception and Case 3 is terminated because the queue is empty and CL≤ N.

| Data Structures used | |
|---|---|
| **struct node** // Structure of a concept in the ontology | **Global variables** |
| Name  String | cNode  structNode  //current Node under investigation |
| Premise  2D Vector // Two dimensional vector with name of premise and its probability value | N  Integer  // Levels of the tree to be investigated |
| Exception  2D Vector // Two dimensional vector with name of exception and its probability value | m, e, k  Real  // parameters of control protocol |
| cNode$\gamma$  Real  // confidence of the decision with 0-degree of strength (if-then relationship) | CL  Integer // current level under investigation |
| cNodeX  Real  //------- | **Functions used** |
| cNode$\delta$  Real  //confidence of the decision with 1-degree of strength (also considering exceptions) | boolean ← checkPremise(cNode) |
| cNode$\omega$  Real  //confidence of the decision with 2-degree of strength (also considering hierarchy) | cNode$\delta$ ← calc$\delta$ (cNode) |
| | cNode$\omega$ ← calc$\omega$ (stack.peek(),cNodePremiseProb,cNode$\delta$) |
| Queue  // To store the nodes to be processed next | N ← calcN(m,e,k,Nm)  //using Formula 4 |
| Stack  // To save the hierarchy as DBA proceeds down the tree | cNodeX ← calcX(m, e, k, N, CL, cNodeX$_m$) //using Formula 5 |

| **Definitions of Functions** | **The DBA Algorithm Stated (DBA)** |
|---|---|
| **checkPremise** | *Input: {root: topmost node of the ontology of type NODE, Context information (CI) =<m,e,k>}* |
| loop in Premises (cNode) | |
| iPremiseName ← fetch name of premise of cNode | |
| iPremiseProb ← Ask the user about the probability of iPremiseName | *Output: answer to input query along with a confidence value* |
| Same iPremiseProb in premise of cNode | *Local Variables: CL: Integer* |
| If (iPremiseProb ≤ DF$_{Thres}$) | 1. CL ← 1 |
| return false | 2. N$_m$ ← fetch maximum number of levels available in the root |
| return true | |
| **calc$\delta$**   //using formula 1 | 3. N ← calcN (m,e,k,N$_m$) |
| cNodeX$_m$ ← fetch the number of exceptions in cNode | 4. Queue.insert(root) |
| cNodeX ← calcX (m, e, k, N, CL, cNodeX$_m$) | 5. while (queue is not empty and CL≤N) |
| cNode$\delta$ = cNode$\gamma$ | 6.   cNode ← queue.removeFirst() |
| loop in the exceptions of cNode till cNodeX | 7.   if (checkPremise (cNode) ) |
| ask user the truth value of exception i | 8.     queue.removeAll() |
| if(exception i is true)  return 1 | 9.     cNode$\delta$ ← calc$\delta$ (cNode) |
| else if (exception i is unknown)  cNode$\delta$ += 0 | 10.    if (cNode$\delta$==1) break |
| else if (exception i is false) cNode$\delta$ += $\delta$ value of exception i | 11.    cNode$\omega$ ← Calculate confidence of |
| return cNode$\delta$ | 12.    current Node using calc$\omega$ |
| **calc$\omega$**   //using formula 2 | 13.    if(cNode$\omega$ ≥ m) |
| iPremiseProb ← Ask the user about the probability of iPremiseName | 14.      CL++ |
| confidence of the root node is 1 | 15.      stack.push(cNode) |
| iConfidence ← fetch confidence of top node | 16.      loop in children of cNode |
| Mp ← find min (iConfidence, iPremiseProb) | 17.        q.insert(child) |
| calc$\omega$ ← Mp × calc$\delta$ | 18. print (pop(stack)) |
| return calc$\omega$ | |

**Complexity of DBA**
Let cNode has 'd' number of defining properties and 'e' number of exceptions.
If N is the total number of nodes of the ontology and 'n' is number of children of a node.
The time (T) and space (S) complexity of the used functions will be as follows:
checkPremise() : T - O(d) and S - O(d)
calc$\delta$(): T - O(e) and S - O(e)
calc$\omega$(): T - O(d) and S - O(d)
DBA: T - O(N×(e+d+n)) and S - O(N)

## 5. Evaluation of the Proposed Approach

This section focuses on the ranking of the human context ontology (SupOnt-NRC), subjective testing, and experimental result and discussion over the diagnostic belief algorithm.

### 5.1. Ranking of SupOnt-NRC

We have incorporated all the essential elements for modeling context in SupOnt-NRC ontology by representing every concept in ontology as a knowledge unit. 'User' element models the information about user's profile like user can be a farmer, teacher, and salesman. 'Time' element models temporal knowledge like day and night. 'Activity' element describes different set of activities like lives in city, works in city, working night shift, doing over time and so on. 'Device' element models the knowledge about devices like harvesting tools etc. 'Service' element models the knowledge about the type of services like 'What is P doing?', 'Is P at home?', 'Is P working outdoor?' or 'Is P planting crops'. 'Location' element describes the location of the context for example, indoor activity or outdoor activity. Each actor is designated his/her respective role. In the current scenario, the role of the user is his current task which is performing right now. This role keeps changing with time. The element 'provider' provides the services according to the choice of the users. 'Environment' element models information about the environment like bad weather. The confidence of the answer determines the certain value of the given answer and reduces the problem of 'uncertainty management'. Semantic context reasoning provides the answer according to the context of the user and maintains the balance between certainty and specificity that shows 'behavior granularity'. 'Proper interface' is available for input/output functions. Server elements within the ontology act as 'context source' like the user, device, available resources, required threshold, and so on. 'Message' element is provided via a list of actions that are recommended after identification of the event.

SupOnt-NRC ontology supports 5Ws theory (Who, When, Where, What, and Why) because the element 'user' defines 'who', the element 'time' defines 'when', the element 'location' defines 'where', the element 'service' defines 'what' and the elements 'activity' and 'device' defines 'why'. We have compared our approach with all the human context ontologies mentioned in Table 2, according to the essential features. Figure 4 shows the overall rating of human context ontologies that has been calculated according to the below-mentioned formula:

$$Overall\ Ranking = \frac{Available\ essential\ features}{Total\ number\ of\ essential\ features}$$

The results show that the overall ranking of SupOnt-NR Ontology is highest as compared to other human context ontologies because it incorporates more number of essential elements that are required to model context.

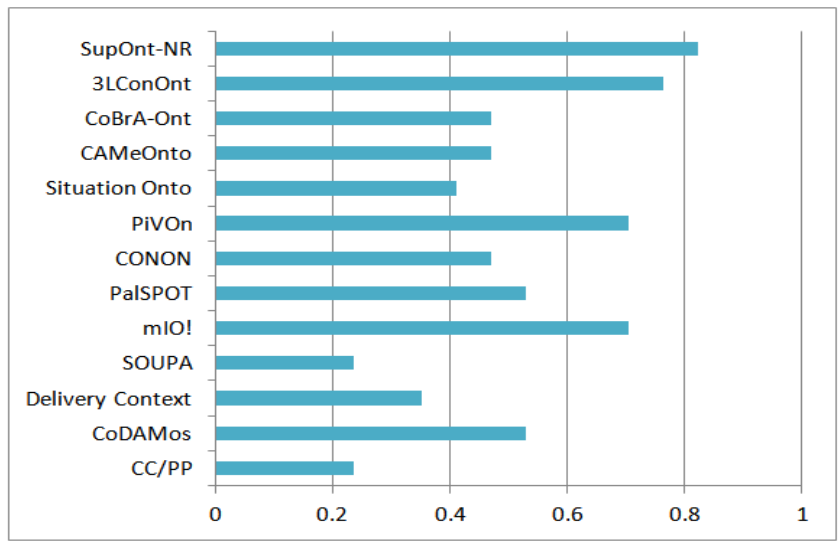

**Fig 4. Overall ranking of human context ontologies**

### 5.2. Subjective Testing

We have conducted subjective testing to gain a deep intuitive understanding of the performance of the proposed approach. Eight categories of questionnaire have been prepared including: (1) Self-efficacy (How well the system can execute courses of action required to deal with a wide array of situations), (2) Usefulness (The degree to which the proposed approach is usable for the end users), (3) Ease of Use (The user-friendliness of the user interface), (4) Response Time (How quickly the proposed algorithm reacts to the posed queries), (5) Relevance (How appropriate and important is the developed system in the current context), (6) Adaptability (Does it adjust to new events, i.e., has the capability to modify itself for changing needs), (7) Acceptability (The extent to which it meets the needs of the end users. Is it able to serve the purpose it is intended for?), (8) Overall (overall ranking of the approach) in order to effectively capture the topic under investigation. All the questions have been validated by experts of the field after a long session of brainstorming.

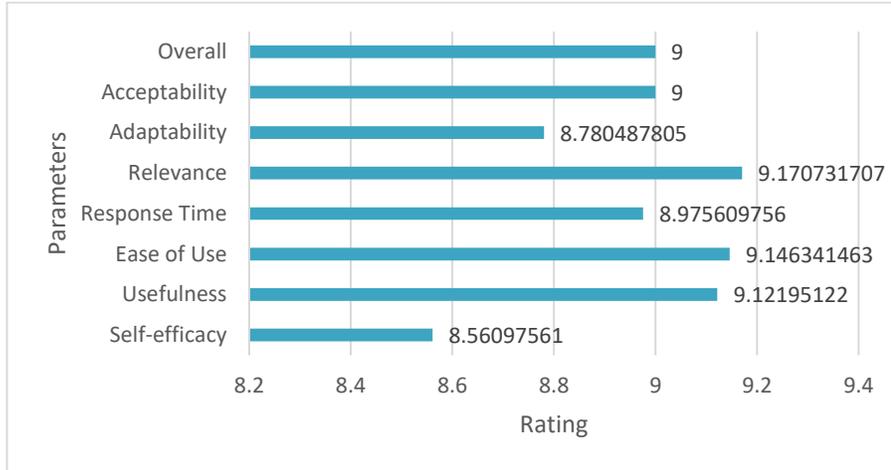

**Fig 5. Subjective Testing**

Forty-one users tested our approach and provided rating from 1 to 10 (1 is minimum and 10 is maximum rating). Among the forty-one users, female and male are 46.3% and 53.7% respectively with median 27 (minimum age 21 and maximum age 52), variance 25.828 and standard deviation 5.082. The participated users are from different occupations like students, faculty, industry, and others. Along with this testing, users also provided written comments on their experience. Users' comments are mostly found positive like quite good, Excellent, Very useful System, Very efficient and so on. The response of the users is collected by google form. Google Form Link: https://bit.ly/342HpM2 , Excel File of the user responses: https://bit.ly/2IFiLsE**.** Figure 5 summarizes the results obtained during the subjective testing. Although the size of the ontology is small but it shows a clear understanding of our proposed work. The parameters Relevance, Ease of Use and Usefulness are the strong points of the proposed approach.

### 5.3. Experimental Results and Discussion

Let the system be asked queries like 'What is P doing?', 'Is P at home?', 'Is P working outdoor?' or 'Is P planting crops'. The input to the DBA will be the root node of SupOnt-NRC ontology and the output will be the answer to his query with a confidence value based on the priority of the user and the context of the user defined by CI = <m,e,k>. As the algorithm proceeds, the user is asked about his confidence in the truthfulness of premises of each node and truth value of each censor. The algorithm is implemented in Java programming language and run over SupOnt-NRC ontology with different input values from each user. Code is available at Github Link: https://bit.ly/3nTgJ8i. The results have been presented below followed by a discussion on findings.

#### 5.3.1. Results

*Assumptions:* We have assumed $DF_{Thers}$=0.88. We have run the DBA as defined in section 4.2 for three different ranges of priority for each of the five priorities of users as defined in table 4, making a total of 15 test cases. The answer of query "where is Robert?" for all the five test cases illustrated below. Each run of the DBA with different user context will require his confidence in the premises of the nodes. We present these DF values for all the nodes of all levels pertaining to person Robert and city Mumbai once in table 5.

**Table 5. Confidence of Premises**

| Level | Premises |
|---|---|
| 1 | livesInCityR:0.90, worksInCityR:0.89 |
| 2 | timeNight: 0.81, timeDay: 0.92 |
| 3 | workingDay: 0.95, daySunday: 0.75 |
| 4 | isFarmer: 0.97, hasSoilFertile:0.98, hasHarvestingTool: 0.55, isTeacher: 0.45, isShopkeeper: 0.67 |

**Test Case1.** Type of User: Very Low Priority User (VLP); Input Context CI = <m=0.3,e=0.125,k=0.125>
    Intermediate Output: N=1, $X_1$=1; Exception at level 1: false

*Output: Robert is in city Mumbai with confidence 0.8544*

**Test Case2.** Type of User: Very Low Priority User (VLP); Input Context CI = <m=0.3, e=0.125, k=1>
Intermediate Output: N=1, $X_1$=1, Exception at level 1: true
*Output: Robert is not in city Mumbai with confidence 1.*

**Test Case3.** Type of User: Very Low Priority User (VLP); Input Context CI = <m=0.3, e=0.125, k=4>
Intermediate Output: N=2, $X_1$=1, $X_2$=1; Exception at level 1: false; Exception at level 2: false
*Output: Robert is outdoor in city Mumbai with confidence 0.803136*

**Test Case4.** Type of User: Low Priority User (LP); Input Context CI = <m=0.4, e=0.6, k=0.125>
Intermediate Output: N=1, X=1; Exception at level 1: false
*Output: Robert is in city Mumbai with confidence 0.8544*

**Test Case5.** Type of User: Low Priority User (LP); Input Context CI = <m=0.4, e=0.6, k=1>
Intermediate Output: N=2, $X_1$=2, $X_2$=1; Exception at level 1: false, false; Exception at level 2: true
*Output: Robert is in city Mumbai with confidence 0.8633*

**Test Case6.** Type of User: Low Priority User (LP); Input Context CI = <m=0.4, e=0.6, k=4>
Intermediate Output: N=4, $X_1$=2, $X_2$=1, $X_3$=1, $X_4$=1; Exception at level 1: false, false; Exception at level 2: false; Exception at level 3: false; Exception at level 4: false
*Output: Robert is planting crops, Workingoutdoor in city Mumbai with confidence 0.716962017*

**Test Case7.** Type of User: Medium Priority User (MP); Input Context CI = <m=0.5, e=1, k=0.125>
Intermediate Output: N=1, X=2; Exception at level 1: false, false
*Output: Robert is in city Mumbai with confidence 0.8633*

**Test Case8.** Type of User: Medium Priority User (MP); Input Context CI = <m=0.5, e=1, k=1>
Intermediate Output: N=2, $X_1$=2, $X_2$=2; Exception at level 1: false, false; Exception at level 2: false, false
*Output: Robert is outdoor in city Mumbai with confidence 0.828768*

**Test Case9.** Type of User: Medium Priority User (MP); Input Context CI = <m=0.5, e=1, k=4>
Intermediate Output: N=4, $X_1$=2, $X_2$=2, $X_3$=2, $X_4$=2; Exception at level 1: false, false; Exception at level 2: false, false; Exception at level 3: false, true
*Output: Robert is outdoor in city Mumbai with confidence 0.828768*

**Test Case10.** Type of User: High Priority User (HP); Input Context CI = < m=0.6, e=2, k=0.125>
Intermediate Output: N=1, X=2; Exception at level 1: false, false
*Output: Robert is in city Mumbai with confidence 0.8633*

**Test Case11.** Type of User: High Priority User (HP); Input Context CI = < m=0.6, e=2, k=1>
Intermediate Output: N=3, $X_1$=3, $X_2$=2, $X_3$=2; Exception at level 1: false, false, false; Exception at level 2: false, false; Exception at level 3: false, false
*Output: Robert is Workingoutdoor in city Mumbai with confidence 0.81202176*

**Test Case12.** Type of User: High Priority User (HP); Input Context CI = < m=0.6, e=2, k=4>
Intermediate Output: N=4, $X_1$=2, $X_2$=2, $X_3$=2, $X_4$=2; Exception at level 1: false, false; Exception at level 2: false, false; Exception at level 3: false, false; Exception at level 4: true
*Output: Robert is Workingoutdoor in city Mumbai with confidence 0.79561728*

**Test Case13.** Type of User: Very High Priority User (VHP); Input Context CI = <m=0.8, e=5, k=0.125>
Intermediate Output: N=2, $X_1$=3, $X_2$=2; Exception at level 1: false, false, false; Exception at level 2: false, true
*Output: Robert is in city Mumbai with confidence 0.8811*

**Test Case14.** Type of User: Very High Priority User (VHP); Input Context CI = < m=0.8, e=5, k=1>
Intermediate Output: N=3, $X_1$=3, $X_2$=3, $X_3$=2; Exception at level 1: false, false, false; Exception at level 2: false, false, false; Exception at level 3: false, false
*Output: Robert is Workingoutdoor in city Mumbai with confidence 0.83739744*

**Test Case15.** Type of User: Very High Priority User (VHP); Input Context CI = <m=0.8, e=5, k=4>
Intermediate Output: N=4, $X_1$=3, $X_2$=3, $X_3$=3, $X_4$=2; Exception at level 1: false, false, false; Exception at level 2: false, false, false; Exception at level 3: false, false, false; Exception at level 4: false, false
*Output: Robert is planting crops, Workingoutdoor in city Mumbai with confidence 0.8290234656*

### 5.3.2. Findings

We have drawn some findings related to the confidence of the answer, threshold, specificity vs. certainty and variation in confidence. Each of these findings are illustrated in Table 6 and discussed below.

**a) Confidence of answer (CF) vs. Threshold (m):** Threshold decides the qualitative expectation of the user for the confidence. If the threshold increases, then the value of CF increases monotonically with same resources irrespective of the value of k. Table 6(a) shows the value of CF for different m and k value where e=1.

**b) Variation in Confidence:** The variation of confidence is calculated according to equation 6.

$$\text{Variation in confidence} = \text{max confidence} - \text{min confidence} \quad (6)$$

The variation of confidence obtained by all five types of users are listed in Table 6 (b). If m increases, then variation in confidence decreases monotonically. So, low priority users achieve high variation in confidence whereas high priority users achieve less variation in confidence. The high value of m generates the same confidence for different combinations of e and k. For example: m=0.75 generates same confidence for different value of e (6,7,8) and k (0.25, 0.5, 1, 2, 4).

**c) Confidence of answer (CF) vs. Specificity (N):** The different aspects of confidence are the certainty of belief, the specificity of conclusion, the available resources, and the allowed threshold. Certainty and specificity are inversely proportional to each other. When a user demands a highly specific answer then the certainty of that answer automatically reduces. For example: if the query is 'where is Robert' then the more certain and less specific answer of this query is 'Robert is in city Mumbai' but the less certain and highly specific answer is 'Robert planting crops in city Mumbai'. The confidence of the answer shows how much answer is certain. Table 6 (c) and (d) shows the variation in CF, N and k when m=0.55, e=0.8 and m=0.75, e=8 respectively. Three majors finding that inferred from table 6 are as follows:

- Irrespective of m and e; if k decrease then CF increases monotonically
- Irrespective of m and e; if k increases then N increases monotonically
- When the N decreases then CF increases monotonically

**d) Exceptions explored (X) vs. Level of tree explored (CL):** Some nodes have exceptions. If exceptions evaluate to be false, then the strength of the decision increases; if true, strength of that node becomes zero; if exception unknown then strength does not change. How many exceptions are evaluated depends on the contextual information (CI). For any value of N, as the current level (CL) increases, the total number of evaluated exceptions (X) at each node monotonically decreases. Table 6 (e) depicts the total number of exceptions (X) that is evaluated at each level (CL).

**Table 6. Finding (a) Confidence of answers vs. Threshold (b) Obtained Confidence (c) Relationship between CF and N when m=0.55, e=0.8 (d) Relationship between certainty and specificity when m=0.75, e=8 (e) Exceptions checked vs. level**

| k | m=0.45 | m=0.47 | m=0.5 | m=0.54 |
|---|--------|--------|-------|--------|
| 0.25 | 0.854400 | 0.854400 | 0.863300 | 0.863300 |
| 0.5 | 0.828234 | 0.828234 | 0.863300 | 0.863300 |
| 1 | 0.770754 | 0.786646 | 0.828768 | 0.828768 |
| 2 | 0.770754 | 0.770754 | 0.795617 | 0.795617 |
| 4 | 0.732216 | 0.755836 | 0.763792 | 0.763792 |

(a)

| Priority of User | m | Variation |
|---|---|---|
| VLP | 0.3 | 0.1448293440 |
| LP | 0.4 | 0.1448293440 |
| MP | 0.5 | 0.1221834720 |
| HP | 0.6 | 0.0995074112 |
| VHP | 0.8 | 0.0168325344 |

(b)

| <m,e,k> | CF | N |
|---|---|---|
| <0.55,0.8,0.25> | 0.863300 | 1 |
| <0.55,0.8,0.50> | 0.863300 | 1 |
| <0.55,0.8,1> | 0.828768 | 2 |
| <0.55,0.8,2> | 0.795612 | 3 |
| <0.55,0.8,4> | 0.763792 | 4 |

(c)

| <m,e,k> | CF | N |
|---|---|---|
| <<0.75,8,0.25> | 0.845856 | 2 |
| <0.75,8,0.50> | 0.837397 | 3 |
| <0.75,8,1> | 0.829023 | 4 |
| <0.75,8,2> | 0.829023 | 4 |
| <0.75,8,4> | 0.803901 | 4 |

(d)

| CI=<m,e,k> | <0.3,0.25,2> | | <0.4,0.6,2> | | | <0.45,1.2,2> | | | | <0.55, 2,2> | | | | <0.75,5,2> | | | |
|---|---|---|---|---|---|---|---|---|---|---|---|---|---|---|---|---|---|
| N | 2 | | 3 | | | 4 | | | | 4 | | | | 4 | | | |
| CL | 1 | 2 | 1 | 2 | 3 | 1 | 2 | 3 | 4 | 1 | 2 | 3 | 4 | 1 | 2 | 3 | 4 |
| X | 1 | 1 | 2 | 1 | 1 | 2 | 2 | 2 | 1 | 3 | 2 | 2 | 2 | 3 | 3 | 3 | 2 |

(e)

## 6. Conclusion

The work presented in this paper addresses the problems of imperfect knowledge in the ontology that arises when representation is not omniscient and proposes a semantic contextual reasoning over the proposed representation of knowledge that provides answer according to the context of the users. Semantic contextual reasoning is discussed by control protocol and diagnostic belief algorithm. It enables the trade-off between the confidence of decision and efforts needed to derive those decisions. The diagnostic belief algorithm describes how a reasoning mechanism depending on the various control parameters (m, e and k) can make a controllable trade-off between the certainty of belief in a conclusion and its specificity. We have shown that representing knowledge as a unit semantically identify the event or activity and provides confidence of the answer which is certain and precise. The results show how confidence is varied according to the context of the user. Some essential features that are required to be incorporated in any context ontology are found to be supported by SupOnt-NRC ontology. The proposed control protocol and the algorithm proved to be logically sound and seem to be a direct consequence of representing knowledge in a manner that is complete. The research presented in this paper can be utilized in various use cases across a multitude of domains demanding for modeling the personalized behavior based on the context of the user.

**Future Work**

In this work, our focus is not on the computation time but to offer a semantic contextual reasoning (which itself is a novel approach by using ontology) that provide answers to the imposed queries according to the user context. In future, we will work on the computation time of merging the domain ontology with upper ontology in order to support all the essential parameters of the Human context ontologies along with the reduction of the time and space complexity of the DBA.